\documentclass{article}

\title{Evolving Accuracy: A Genetic Algorithm to Improve Election Night Forecasts}
\author{Ronald Hochreiter and Christoph Waldhauser}
\date{October 2012}

\usepackage{geometry}
\usepackage{clrscode}
\usepackage{url}
\usepackage{graphicx}
\usepackage{natbib}

\begin{document}

\maketitle

\begin{abstract}
\noindent In this paper, we apply genetic algorithms to the field of electoral studies.
Forecasting election results is one of the most exciting and demanding tasks
in the area of market research, especially due to the fact that decisions have
to be made within seconds on live television. We show that the proposed method
outperforms currently applied approaches and thereby provide an argument
to tighten the intersection between computer science and social science, 
especially political science, further. We scrutinize the performance
of our algorithm's runtime behavior to evaluate its applicability in
the field. Numerical results with real data from a 
local election in the Austrian province of Styria from 2010 substantiate the 
applicability of the proposed approach.
\end{abstract}
\section{Introduction}\label{introduction}

When the last ballots have been cast and the last polling station
closes, the fruits of a stressful afternoon are brought to bear: the
first election forecast is being broadcast over the air. Much of the
work behind it, however, actually took place long before that, starting
weeks before the election and culminating shortly after noon.

Forecasting elections is arguably the most demanding and stressful but
also the most exciting task market researchers can perform
\citep{fienberg2007memories}. The term election forecast can mean
different things. \citet{karandikar2002predicting} and
\citet{morton1990election} give a summary of different meanings, and
when we speak of election forecasting in this paper, we mean exclusively
what they termed a \textit{results-based forecast}, a forecast based on
partially counted votes without any exterior information like polls or
surveys.\footnote{Contrast this with an approach that uses external,
  here historical, data in \citet{chen2007dynamic}.} In the traditional
forecasting process, after weeks of preparation, decisions have to be
made in split seconds, possibly on live television. The preparation in
the weeks before the election is a tedious process that involves many
person hours and is error prone. This paper improves the current
situation of the industry by contributing solutions based on genetic
algorithms\footnote{Preliminary results of this research project were
  presented at ACM GECCO conference 2011 \citep{hochreiter2011evolved}.}
to the most expensive and fragile elements of the field.

The remainder of this paper is structured as follows. First we will give
an introduction to the methodology that constitutes the foundation of
industry standard election forecasting as it is practiced today. Then
the elements of this forecasting process that are especially expensive
and fault intolerant are identified. In Section
\ref{optimization-process} we describe how genetic algorithms can be
used to find near optimal solutions to the problems identified above.
The devised algorithm is described in detail and evaluated using a
standard set of indicators and real data from the field. Results of this
analysis are presented in Section \ref{results}. Finally, we offer some
concluding remarks and suggestions for further developments.

\subsection{Ecological Regression}\label{ecological-regression}

Forecasting elections is a business that depends on meticulous
preparations and accurate knowledge of the political processes behind
the scenes. In the beginning of televised live election night
forecasting, sometimes disastrous miscalculations paved the way for
numerous endeavors, that were undertaken to improve the status quo
\citep{mughan1987general, morton1988election}. Today, election
forecasting using a methodology termed ecological regression
\citep{fule1994estimating, brown1999forecasting, hofinger2002orakel, king1997solution, king2004ecological, greben2006model},
engages in the daunting task of comparing polling stations or
constituencies of a geographical entity from a past election\footnote{To
  make things worse, past election results do not easily translate into
  new election results because of old people dying and young ones
  becoming eligible to vote. Assuming, admittedly somewhat naively, that
  new voters behave in general similar to old voters, this transition
  becomes merely an exercise in multiplying old vote shares with the
  number of new voters. Any deviance in voting behavior will be
  accounted for in the regression model introduced later.} to the
present one that is meant to be forecast\footnote{The election
  forecasting framework used in this paper is but one of many. Other
  frameworks are used to forecast elections well before they take place,
  usually with the aim of only predicting the winner, and not producing
  precise estimates for vote shares
  \citep{whiteley2010aggregate, fisher2010polls, visser1996mail, sanders1995forecasting}}.
This only works because (1) not all polling stations provide their
results at the same time and (2) voters that go to one polling station
will behave similar to voters at another one.

Let's look at the election forecasting methodology in a little more
detail. In a multi party system for any given election there are
multiple parties competing against each other for votes. Voters can cast
these votes at polling stations which are usually located close to their
homes. It is also clear, that at least for developed democracies,
parties have a history of performances in past elections. Any election
forecast uses (at least) two elections, one in the past and the current
one with the overall aim of predicting the vote shares of the current
one. Since voters have formed an opinion and elect a party
accordingly\footnote{For two competing theories of how this might
  happen, see \citet{lau2006voters} and \citet{lewis2008american}.}, not
all parties will end up with the same share of votes, when comparing two
elections.

Note that there are two different kinds of vote shares that can be used
as performance metrics for parties: either the proportion of the total
electorate voting for a party or the proportion of the constituency that
actually did cast a valid vote, that voted for a party. In the
following, these quantities are called \%Elec and \%Vald, respectively.
Most clients will be interested in the latter one, as it constitutes the
post-election political reality.

In this model the performance $p$ of a party $i$ at a current election
is a linear combination of all $j$ parties' performances at the
reference election plus the proportion of nonvoters (NV), for all $k$
polling stations. To simplify things, the nonvoters are considered to be
just another ordinary party and are thus included in the $j$ parties. So
the following equation has to be estimated for all $j$ parties to link
the old election results from polling station $k$ to the new election
results at that polling station:

\begin{equation}
p_{i,k} = \sum_j x_{j,k} p_{j,k} + p_{NV,k}
\end{equation}

The factor $x_k$ in the equation above is the quantity of interest in
the election forecasting process. This quantity can be considered as a
transition multiplier. For instance a value of $x_{n,k}=0.6$ for two
parties $m,n$ means that in the current election party $m$ could
mobilize 60 percent of the last time voters of party $n$ for its own
cause at polling station $k$. If $m=n$, $x_{m,k}$ boils down to the
proportion of traditional $m$ voters the party could again re-win at the
current election (at this polling station); for all $m \neq n$, the
different $x_{n,k}$ sum up to the votes that were won by party $m$ from
competing parties. All $x_k$ of a polling station $k$ together make up a
matrix with as many rows as parties in the current election and columns
as parties in the old election. This matrix projects the old election's
vote shares into the space of the new election. In the most trivial
example, the same, let's say 4 parties compete in both elections. This
means that the equation from above needs to be estimated four times for
each polling station, leading to a $4\times 4$ projection matrix for
each polling station.

Obviously, a projection matrix can only be established for polling
stations that already reported their results. The polling stations that
did not yet report their results are then to be forecast. As stated
earlier, it is assumed that any trend visible from the already declared
polling stations will also apply to the polling stations not yet
counted. So the idea is now to use the already obtained projection
matrices on the old election results from those polling stations still
missing. When, and this is quite quickly happening during an election
day, more than one polling station have their results reported multiple
projection matrices will be available. Then a cell-based average
function over the available projection matrices is used to obtain an
overall matrix.

Unfortunately, not all polling stations will follow the general trend,
or will follow it only to some extend. Therefore, care must be taken in
choosing the projection matrices that are used as input in computing the
overall matrix.

As stated above, this method relies on the assumption, that voters will
behave similarly. However, consider that on the math side of things, as
the used regression models are unbounded, this method is a linear
approximation of the choices the electorate makes. Also, since
regression can be considered as computing an average over a number of
data points, $x$ can take extreme values if there are heterogeneous
trends between polling stations. This poses a problem to the election
forecasting model as percentages below $0$ and above $100$ can not be
accounted for by voter mobilization.

The solution to this problem lies in grouping polling stations together
that will exhibit a similar trend in the transition from the reference
election to the current election. So the mean of the individual
projection matrices are computed only for a subset of the available
matrices.

So to summarize, in the election forecasting process, the relationships
between old and new party results are used to project the results for
yet missing polling stations. By means of multiple regression models,
the transition multipliers are estimated per polling station. The
transition multipliers of similar polling stations are then combined
into averages. These averages are then used to compute the votes the
parties are likely to obtain in the missing locations.

Traditionally the grouping, the identifying of polling stations that
will exhibit similar trends, is done by experienced senior researchers
using K-means clustering (see \citet{MacQueen1967}) and constant size
binning techniques. This process is usually very time consuming (and
thus expensive), as there are no fixed rules and many different
possibilities have to be evaluated by hand. Additionally, there is no
guarantee that the groupings found in such a way will actually be
homogeneous. Given the small number of possible combinations that can be
tried in manual assessment, they are even quite unlikely to be related
at all. If the resulting groups of polling stations, however, are
homogeneous enough, stable forecasts will be available at a very early
state of the vote counting process.

To summarize, for any election forecasting endeavor using the
aforementioned method, the grouping of polling stations into homogeneous
clusters is crucial. The search for a perfect grouping is a tedious and
time consuming process especially given the huge number of possible
combinations.

\section{Optimization Process}\label{optimization-process}

In this section we will describe the genetic optimization procedure we
used to improve the quality of the grouping solutions and thus the
quality of election forecasts. We will first cast a closer look at the
groupings of polling stations and ways on assessing the quality of an
election forecast. Then we will present the pseudo code of the genetic
algorithm used for the optimization process.

\subsection{Opportunities for
Optimization}\label{opportunities-for-optimization}

When grouping polling stations together into homogeneous groups, some
issues have to be considered:

\begin{itemize}
\itemsep1pt\parskip0pt\parsep0pt
\item
  Groups should exceed a minimum size.
\item
  Each group needs to contain polling stations that will be declared
  early in the race.
\item
  Overly large or small groups should be avoided.
\item
  Being able to attribute external meaning to the grouping aids in
  interpretation.
\end{itemize}

The method used in election forecasting is based on regression. As
regression becomes computationally unstable when too few data points are
available, small groups are a hazard to the computation. It is also
important to consider that only a fraction of the polling stations in
each group will actually be available to estimate the regression
coefficients (ie, the transition multipliers) early on election night.
Conversely, overly large groups are as well problematic, as large groups
are often typical of cluster algorithms which thereby fail to detect
structure in the data. In this case, one large,
\textit{average, typical} group is found, with only a few outliers
appertaining to the other groups. Finally, it can be helpful for an
interpretation of the election results to be able to describe the groups
and thus characterize the environments of the voters causing a certain
trend.

For the optimization problem at hand we used publicly available
data\footnote{Election results of Austria are available from the website
  of the Austrian Federal Ministry of the Interior at
  \url{http://www.bmi.gv.at/cms/bmi_wahlen/}.} on a local election in
the Austrian province of Styria from 2010. We used the Styrian part of
Austria's general election of 2008 to predict the outcome. The data
available is polling station data aggregated to the constituency level,
resulting in $542$ data points. There were seven parties competing,
eight including nonvoters.

The selected election data is typical for the industry in that it is
rather difficult to predict. Styria consists of mostly rural communities
in secluded alpine valleys with a few larger cities in between. The
largest share of the votes, however, originates in the provincial
capital of Graz. As rural and urban areas have very little in common
regarding voting behavior, the prediction becomes tricky.

To assure a sensible grouping, the data points were split into ten
groups initially. This would yield approximately $54$ constituencies per
group, giving ample data points for an early estimation effort. In a
simulated election forecast it was pretended that a part of the
constituencies had not yet been declared. This part amounted to $90$
percent of the entire electorate, roughly spread out over $36.9$ percent
of the constituencies.

The quality of an election forecast was established by considering the
root mean squared error (RMSE) of the forecast with respect to the
actually observed election outcome. The exact quantity that was used to
measure the deviation was varied. Details will be given in Section
\ref{sec:Results}.

RMSE was used in spite of \citet{armstrong2001principles} arguing
against it. His main critique is the poor performance of RMSE as an
indicator in forecasting long-run time series data and its sensitivity
to outliers. While this is well founded, it does not apply to the
election forecasting problem. Here, the shortest time series possible is
used. Furthermore sensitivity to outliers is an asset, since clients and
the television audience will be sensitive to them as well.

\% which part we optimized, and why this can work. ie: the clustering \%
solution, consider how many possible combinations there are, justify \%
why we use only 10 groups (ie need to maintain stable regression) \% and
Also justify the use of RMSE in light of \%
armstrong2001principles\ldots{}

\subsection{Genetic Optimization}\label{genetic-optimization}

To optimize groupings in an election forecast, genetic algorithms can be
used. The idea of any genetic algorithm is to start with a number of
random solutions. The best of these random solutions are then combined
together and combined with fresh randomness, so to speak. In a next
step, these children solutions are recombined once more. These steps are
repeated for a very large number of times until they converge towards a
stable and near optimal solution. The mechanics of genetic algorithms
are closely modeled after evolution on a genetic level as observed
throughout nature. Genetic algorithms are generally considered to
provide excellent results on a wide range of optimization
problems\footnote{See \citet{oduguwa2005evolutionary} and
  \citet{karaboga2011novel} for applications to clustering.}.

At the core of a genetic algorithm lie chromosomes, as in natural
genetics as well. Each of these chromosomes reflects a particular
solution to the optimization problem at hand. Each of these chromosomes,
or solutions, is being evaluated at the hands of a pre-defined target
function, for instance the deviance between predicted and observed vote
shares. In this example, a chromosome is the total grouping structure
for all constituencies. Chromosomes are made up of genes, which
represent the group membership of individual constituencies. The closer
a candidate solution, a chromosome, gets to reality as observed, the
better it performs. By recombining the genes of two chromosomes a child
is produced, and an improvement with respect to the target function is
aspired.

Often enough, this simple recombination of parent DNA leads into a dead
end. A dead end, in terms of operations research is a local optimum of a
function. Think of an optimization problem as the search for the highest
peak in an unknown mountainous region. One way to find this peak, is to
climb up until there are no rocks left to climb, on every side there are
only descents to find. We thus have found a peak. However, perhaps it is
not the highest peak, perhaps we need to descend into a valley to climb
a yet higher peak. If that is the case, conventional peak search is at
an end, since all options correspond only to descents. We thus need to
climb down in hopes of finding a higher peak some other place. In terms
of genetic algorithms, fresh genetic information is introduced by means
of mutation and re-seeding, the introduction of totally random
chromosomes into the population to allow for a fresh start that might
lead to an even higher peak.

The algorithm\footnote{The algorithm was implemented using R
  \citep{rcran}; all plots were produced with ggplot2
  \citep{wickham2009ggplot2}.} we found most suitable for the problem at
hand was initialized with a set of random solutions and progressed using
a number of genetic operators.

The optimization problem can be broken down to a clustering problem with
additional constraints described above. We solve this optimization
problem by adapting a standard genetic algorithm, e.g.~as surveyed by
\citet{BlumRoli2003}, and summarized in Table \ref{tab:ea}. The
chromosomes of the algorithm are different grouping solutions, with each
constituency being represented by one gene per solution. A gene
expresses a constituency's group membership.

This adapted algorithm uses three genetic operators: random re-seeding
of populations, one and two point crossovers and mutation. Additionally,
the algorithm adheres to the principles of elitism and elite mixture
breeding. The random re-seeding of the population serves to keep the
gene pool fresh with alternatives to stay out of local optima. This is
done by forcing a part of a new generation's chromosomes to be totally
random.

The crossover operator combines two parent chromosomes into one child
chromosome, cutting randomly at one or two points. Because of elite
mixture breeding, at least one parent comes from the the top performing
solutions, while the other parent is selected from a larger pool.
Mutation is implemented by randomly changing genes in a chromosome. Each
gene has the same probability of being mutated. Each new generation
consists of a share of the top performing chromosomes of the old
generations, a proportion of entirely random chromosomes and the
remainder being offspring produced as described above. Table
\ref{tab:param} gives the proportions and probabilities for the
parameters and operators, respectively. These have been established
experimentally, with the constraints denoted above in mind.

\begin{table}
 \centering
\caption{Parameters of Genetic Optimization}
 \label{tab:param}
 \begin{tabular}{c|l} 
 Parameter&Value\\ \hline
 Initial population size & 100\\
 Generations & 500\\
 Elite proportion & 0.1\\
 Reproduction eligible population proportion & 0.7\\
 Mutation probability & 0.003\\
 Random re-seeding proportion & 0.1\\
 \end{tabular}
 \end{table}

\begin{figure*}
\centering
\includegraphics[height=4in, width=6in]{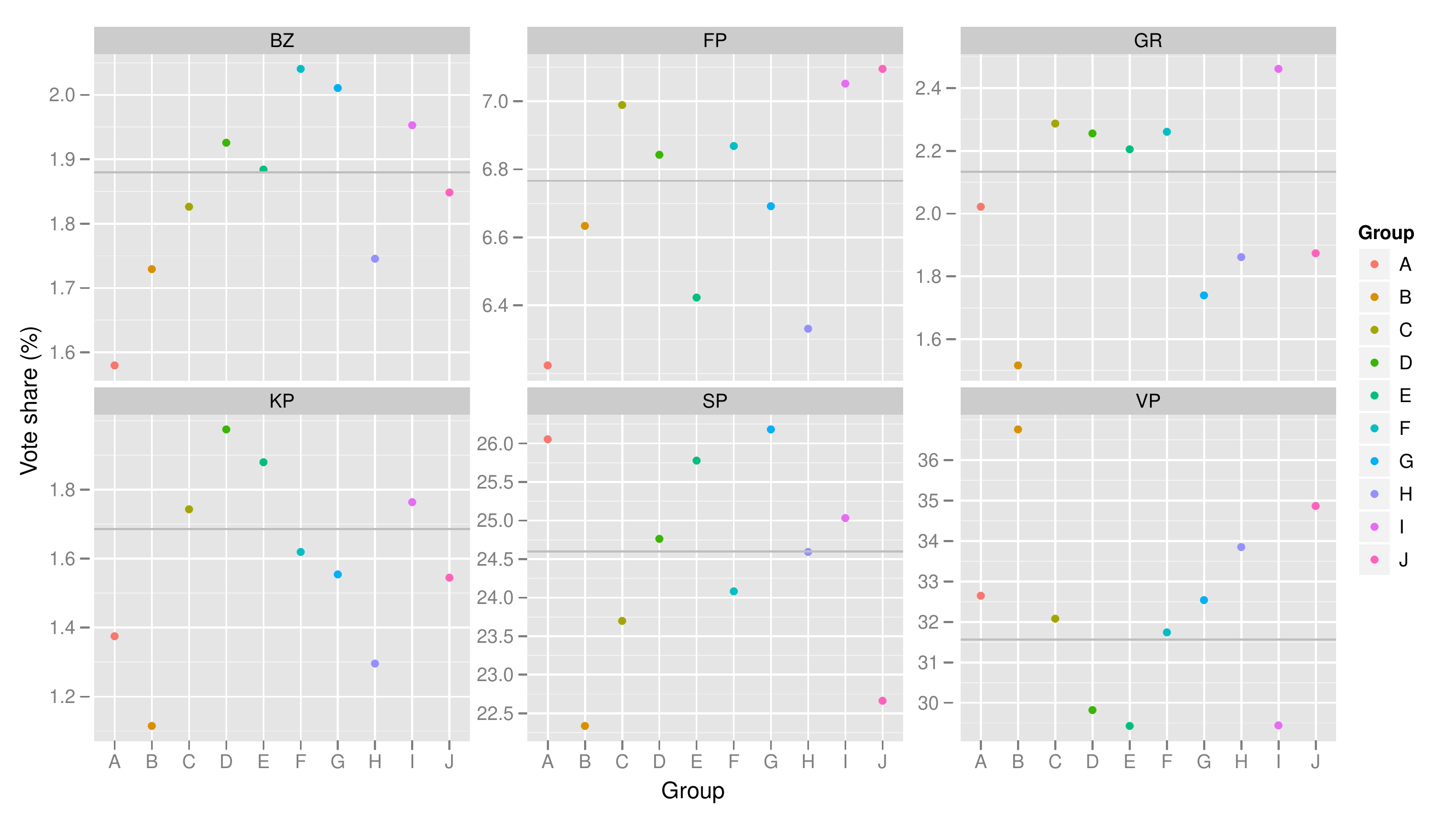}
\caption{Grouping's characteristics with respect to party
  performances, one party per panel. The horizontal line represents that party's mean
  performance across Styria.}
\label{fig:GroupNPlot}
\end{figure*}

\begin{table}
\caption{Meta-heuristic: Genetic Algorithm}
\begin{codebox}
\li $P \gets GenerateInitialPopulation$
\li $Evaluate(P)$
\li \kw{while} termination conditions not met \kw{do}
\li \> $P' \gets Recombine(P)$ 
\li \> $P'' \gets Mutate(P')$ 
\li \> $Evaluate(P'')$ 
\li \> $P \gets Select(P \cup P'')$
\li \kw{end while} 
\end{codebox}
\label{tab:ea}
\end{table}

\% also describe the resulting grouping here, only briefly. At $500$
generations, the algorithm produced a near optimal solution (see Section
\ref{quality-of-the-solution}). The characteristics of this solution
with respect to the competing parties' vote share in the groups are
pictured in Figure \ref{fig:GroupNPlot}. The optimized solution
describes groups that are most homogeneous in their voting behavior;
hence for instance Group B has above average votes for the Conservatives
(VP) and below average votes for left wing parties (SP, GR). Group A is
even more sharply discriminated: here only the Social Democrats (SP)
achieve large above average results. Not all groups, however,
concentrate on partisan logic. Group C, for instance, exhibits an above
average vote share for the Greens (GR). At the same time, also the right
wing party FP is overly popular in this group. This is an indication
that there is a similar trend in those constituencies that prefer small
parties over large ones, protesting the political establishment. Yet
other groups, like H, do not seem to follow any pattern that can be
captured by examining party vote share in this group.

\section{Results}\label{results}

\label{sec:Results}

The performance of the algorithm described above was analyzed with
respect to a number of criteria. First, the influence of the genetic
operands was assessed by considering convergence given their use.
Secondly, the variability of convergence was scrutinized. Finally we
appraised the quality of the solution, given different target functions.

We chose from three different target functions, all returning RMSE
between forecast and true result, with respect to (a) absolute votes,
(b) \%Elec, and (c) \%Vald. All three have their own merits. Minimizing
deviation with respect to percent of the electorate will give fairly
accurate estimations of voter turnout, as it treats nonvoters and
parties alike. This is a feature of interest especially for the
political science community in academia. Industry clients are more
interested in percent of valid votes, as indicated above. Finally,
deviation in absolute votes represents a compromise between both
quantities, so for most of the evaluations below this quantity was used.

\subsection{Effect of Genetic
Operators}\label{effect-of-genetic-operators}

\begin{figure*}
\centering
\includegraphics[height=4in, width=6in]{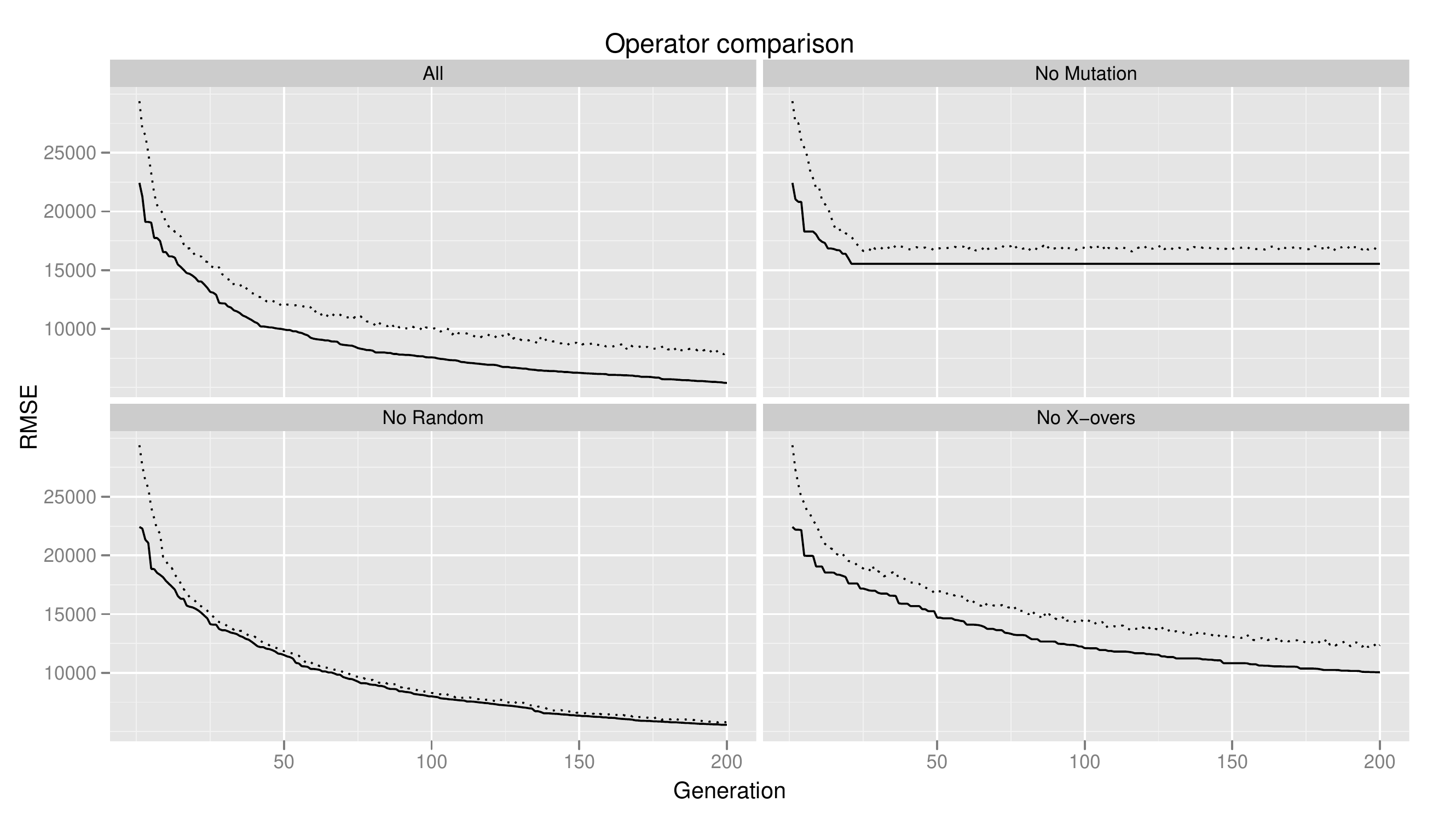}
\caption{Evaluation of genetic operator significance. The dotted line
  represents the generational mean, the solid a generation's best performance, respectively.}
\label{fig:operEval}
\end{figure*}

Each of the three core genetic operators was tested on its significance.
For each test one genetic operator was disabled and the optimization
allowed to evolve over a period of $200$ generations. The results are
shown in Figure \ref{fig:operEval}. The largest impact has the mutation
operator. Without it, the optimization got caught in a local optimum
after as little as twenty generations. Also quite important are the
crossover operators. Without them, the optimization converges at a much
slower pace. The presence of the random genetic operator seems to be
quite negligible. It does have an influence on the variability of the
produced solutions per generation, but this influence seems to have no
effect on the quality of the best solution attained.

\subsection{Convergence Variability}\label{convergence-variability}

\begin{table}
 \centering
 \caption{Convergence of Optimization (Generational Average)}
 \label{tab:convOptim}
 \begin{tabular}{r|l|l} 
 Indicator&Mean&Best\\ \hline
 Min&6382&3683\\
 Median&8024&5421\\
 Mean&9233&6624\\
 Max&28950&21680\\
 St.Dev.&3448&3341\\ 
 \end{tabular}
 \end{table}

The optimization process converges rather quickly\footnote{On the
  evaluation of genetic algorithms see \citet{defalco2002mutation}.}. As
shown above, after only fifty generations, RMSE is reduced to $25$
percent of its initial value. To assess the long-term variability of the
genetic optimization, we collected the results from $10$ optimization
runs, $500$ generations each. The results are presented in Table
\ref{tab:convOptim}. It is clearly visible that the distribution is
highly skewed for both generational means and champions alike.

\begin{figure*}
\centering
\includegraphics[ height=4in, width=6in]{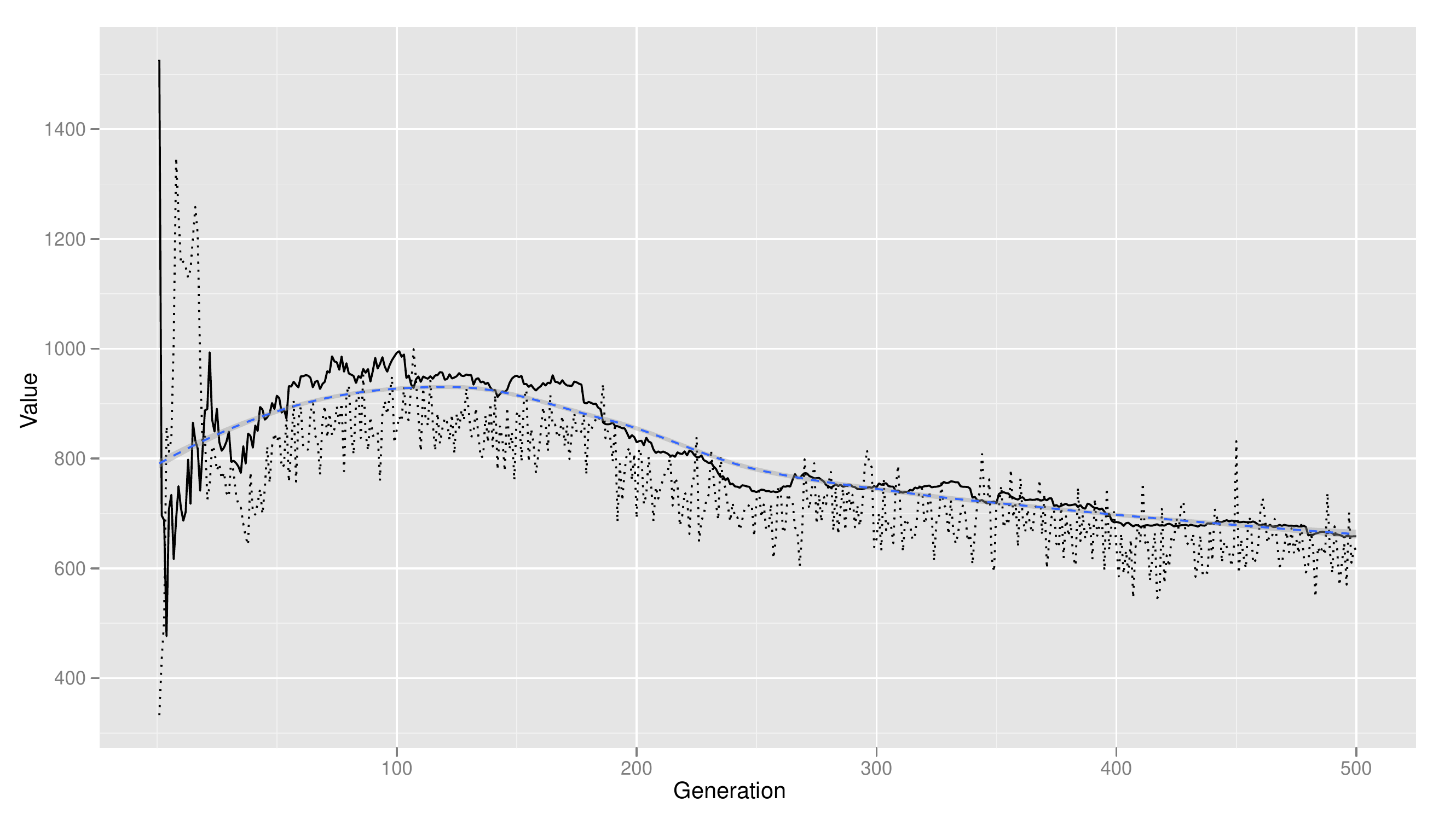}
\caption{Standard deviation (in absolute votes) of best and mean
  performances per generation across $10$
  simulation runs. The dotted line represents the standard deviation
  of the generational means, the solid the
  best performances', respectively. The dashed (blue) line is a kernel density
  estimator of the best performances' standard deviations.}
\label{fig:convSD}
\end{figure*}

In Figure \ref{fig:convSD} the standard deviation for each generation's
best and mean performance is shown. After a turbulent start, variance
constantly increases until generation $102$. After that an almost
monotone decline begins. This can be interpreted as an increase in
solution stability that almost voids any value of a search beyond 250
generations.

\subsection{Quality of the Solution}\label{quality-of-the-solution}

In order to gauge the quality of the proposed algorithm, it was applied
to minimize two different target functions, pertaining to the Absolute
and \%Vald ways of expressing voting results. The resulting grouping
solution was employed in a simulated election forecast. The deviation
between forecast and true result was measured in \%Elec and \%Vald and
is summarized in Tables \ref{tab:devElec} and \ref{tab:devVald}. The
performance attained with a manual grouping solution, the industry
standard to date, is provided for comparison.

It is manifest, that---dependent on the target of the optimization
process---humans are outperformed either in the electorate or valid
votes metric. Also note that optimizing with respect to \%Vald leads to
near perfect predictions.

\begin{table}
 \centering
 \caption{Deviations of Optimized Solutions (\%Elec)}
 \label{tab:devElec}
 \begin{tabular}{r|l|l|l}
 Indicator&Human&OptAbs&OptVald\\ \hline
 Median&0.700&0.550&0.850\\
 Mean&2.742&0.810&3.992\\
 Max&11.600&3.700&17.400\\
 St.Dev.&4.277&1.257&6.655\\ 
 \end{tabular}
 \end{table}

\begin{table}
 \centering
 \caption{Deviations of Optimized Solutions (\%Vald)}
 \label{tab:devVald}
 \begin{tabular}{r|l|l|l}
 Indicator&Human&OptAbs&OptVald\\ \hline
 Median&0.200&0.500&0.000\\
 Mean&0.014&0.029&0.000\\
 Max&0.800&1.300&0.100\\
 St.Dev.&0.537&0.879&0.058\\ 
 \end{tabular}
 \end{table}

\section{Discussion}\label{discussion}

Predicting the outcome of a ballot on election night depends on the
usability of the obtained groupings of the constituencies. We have
proposed a way of improving and vastly surpassing manually derived
groupings by means of a genetic algorithm.

The described algorithm, however, needs to be adapted before it can be
used during election night forecasts. The current runtime for $500$
generations is too long. A pointer towards a solution is our analysis of
the long run performance of the algorithm and the significance of the
genetic operators. While variance in a generation's chromosomes remains
high throughout the optimization process, improvement of the best
solution rapidly degrades after approximately $250$ generations. It can
thus be argued that a halving of prescribed generations or a softer
formulation of stopping criteria in the algorithm would lead to quicker
results that still are significantly better than human groupings.

Another line of research might be interested in the groupings
themselves. From a mathematical point of view, the groupings reflect
most homogeneous partitions of the constituent space. While our
algorithm can find these groups easily, it is up to social science to
explain why the constituents of a group vote in a similar fashion. This
might lead the way to a radically new understanding of the mechanics
behind vote preference emergence.

The logical extension of this paper in the technical realm is the
improvement of the real world deployability of genetic algorithms in the
field of election forecasting. By using distributed computing
environments that are already available for R \citep{rsnow}, genetic
optimization can be used during election night presentations to improve
results at an early stage. While the overall result in this use case is
not yet known, the target function needs to be modified to optimize the
forecast for single polling stations as soon as they are being declared.
This constant optimization requires a considerable amount of processing
power, but is already well within the capabilities of affordable data
center solutions. Ideally, any such system would be complemented by an
optimization algorithm that can handle dynamically changing data, like
the one put forth in \citet{hochreiter2012reva}.

The introduction of genetic algorithms by computer scientists to the
realm of social scientists is akin to crossing into uncharted
territory---for both sides. Despite the numerous obstacles raised by
different communication cultures and epistemological propositions, it is
an endeavor worthwhile. We hope that this paper is a contribution to
building a bridge between what once had been termed incommensurable.
\bibliography{hw-evoaccu}
\bibliographystyle{plainnat}

\end{document}